\documentclass{article}

 \usepackage{amsmath}
\usepackage{arxiv}
\usepackage[utf8]{inputenc} 
\usepackage[T1]{fontenc}    
\usepackage{hyperref}       
\usepackage{url}            
\usepackage{booktabs}       
\usepackage{amsfonts}       
\usepackage{nicefrac}       
\usepackage{microtype}      
\usepackage{lipsum}		
\usepackage{graphicx}
\usepackage{natbib}
\usepackage{doi}
\usepackage{tikz}
\usepackage[table,xcdraw]{xcolor}
\usepackage{colortbl}
\usepackage{graphicx}
\usetikzlibrary{arrows.meta, positioning, shapes.multipart, fit, calc}

\title{Learning to Prioritize IT Tickets: A Comparative Evaluation of Embedding-based Approaches and Fine-Tuned Transformer Models}

\author{%
  \href{https://orcid.org/0000-0002-2233-6280}{\includegraphics[scale=0.06]{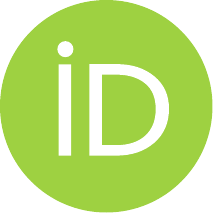}\hspace{1mm}Minh Tri LÊ}\\
  Global AI Lab\\
  EasyVista\\
  \texttt{mle@easyvista.com}
  \And
  \href{https://orcid.org/0009-0004-2715-7363}
  {\includegraphics[scale=0.06]{orcid.pdf}\hspace{1mm}Ali AIT BACHIR} \\
  Global AI Lab\\
  EasyVista\\
  \texttt{aaitbachir@easyvista.com}
}

\renewcommand{\headeright}{}
\renewcommand{\undertitle}{}
\renewcommand{\shorttitle}{Prioritizing IT Tickets: Embedding Clustering and Fine-Tuned Transformer Models}

\hypersetup{
pdftitle={Learning to Prioritize IT Tickets: A Comparative Evaluation of Embedding-based Clustering and Fine-Tuned Transformer Models},
pdfsubject={AIOps, embedding, fine-tuning, transformer, prioritization},
pdfauthor={Minh Tri LÊ},
pdfkeywords={AIOps, embedding, fine-tuning, transformer, prioritization},
}

\begin{document}
\maketitle

\begin{abstract}
Prioritizing service tickets in IT Service Management (ITSM) is critical for operational efficiency but remains challenging due to noisy textual inputs, subjective writing styles, and pronounced class imbalance. 
We evaluate two families of approaches for ticket prioritization: embedding-based pipelines that combine dimensionality reduction, clustering, and classical classifiers, and a fine-tuned multilingual transformer that processes both textual and numerical features. 
Embedding-based methods exhibit limited generalization across a wide range of thirty configurations, with clustering failing to uncover meaningful structures and supervised models highly sensitive to embedding quality.
In contrast, the proposed transformer model achieves substantially higher performance, with an average F1-score of 78.5\% and weighted Cohen’s kappa values of nearly 0.80, indicating strong alignment with true labels. These results highlight the limitations of generic embeddings for ITSM data and demonstrate the effectiveness of domain-adapted transformer architectures for operational ticket prioritization.

\end{abstract}


\section{Introduction}

In today's digital enterprises, IT Service Management (ITSM) platforms generate a continuously growing volume of user-submitted support tickets. These tickets are often expressed in natural language, unstructured, and written in a variety of styles, languages, and technical registers. Managing such heterogeneous data manually is not only labor-intensive but also introduces human bias, delays, and inefficiencies that impact both service quality and user satisfaction. Prioritization, deciding which issues to address first, is a critical aspect of IT support, especially in large-scale environments where operational decisions must balance urgency and business impact in real time.

Recent advances in Natural Language Processing (NLP) offer promising avenues for automating both the categorization \citep{mohanna2025centroidbasedframeworktext} and prioritization of IT tickets. However, as several studies have shown, off-the-shelf machine learning approaches struggle to adapt to the IT ticket domain. Traditional models, such as those based on TF-IDF or Word2Vec, are limited in their ability to understand sentence-level semantics and contextual relationships, particularly in noisy, domain-specific, and multilingual datasets (~\cite{do2021};~\cite{fuchs2022};~\cite{daCosta2023}). Moreover, classical machine learning methods lack the ability to jointly process text with numerical features, such as impact or urgency scores, which are essential for effective prioritization.

While pre-trained language models like BERT and XLNet have demonstrated improved performance on general NLP tasks, their application in ITSM scenarios remains challenging. As noted by~\cite{ZANGARI2023}, even advanced deep learning techniques yield F1-scores barely exceeding 50\% when applied to IT ticket data. It is largely due to the specialized jargon, ambiguous labels, and inconsistent structure found in real world support logs. Furthermore, Cohen’s Kappa, a common metric used to assess classification reliability (~\cite{cohen1968};~\cite{datatab2024}), tends to be sensitive to class imbalances, which is particularly problematic for prioritization tasks where low-frequency but high-criticality tickets are underrepresented.

To address these challenges, we launched a research and development initiative with two complementary objectives: (1) to develop a robust automatic categorization system capable of suggesting relevant IT ticket categories based on free-text input, and (2) to design a prioritization engine that predicts operational impact and urgency to rank tickets by criticality. This second objective, prioritization, is the focus of our paper.

The prioritization task requires sophisticated modeling techniques that can integrate semantic understanding of ticket descriptions with structured indicators of business risk. This includes handling noisy, multilingual input; combining textual and numerical data; and addressing unbalanced class distributions. Moreover, the system must operate under real-time constraints and scale across a wide range of ticket types and service domains.

In this paper, we explore hybrid architectures that combine contextual embeddings from pre-trained language models with numerical features and domain-specific adaptations. We also evaluate prioritization accuracy using multiple performance metrics, including weighted Cohen’s Kappa, to better reflect operational needs in imbalanced scenarios. By investigating domain-specific adaptations of modern NLP techniques, our goal is to bridge the performance gap observed in previous research and provide a viable, production-ready solution for automated IT ticket prioritization.

The remainder of this paper is organized as follows. In Section~\ref{sec:state-of-art}, we present a review of the state of the art, highlighting the main approaches and their limitations in the context of IT ticket prioritization. Section~\ref{sec:headings} outlines our main contributions and the proposed methodology. The experimental setup, including data, preprocessing, and implementation details, is described in Section~\ref{sec:experimental-setup}. In Section~\ref{sec:results}, we report and discuss the results of our evaluation. Finally, Section~\ref{sec:conclusion} concludes the paper and outlines directions for future research.

\section{Related works}
\label{sec:state-of-art}

Automating ticket classification and prioritization in IT Service Management (ITSM) systems has been the subject of growing research attention. Early studies such as~\cite{do2021} and~\cite{fuchs2022} relied on classical machine learning methods with static word embeddings like TF-IDF or Word2Vec. These approaches, while interpretable, are limited in their ability to model context and semantics, particularly for noisy, multilingual ticket data. Moreover, they cannot process structured and unstructured inputs jointly, which is essential when impact or urgency scores must be considered alongside free-text descriptions.

Recent surveys (e.g.,~\cite{daCosta2023}) emphasize the benefits of deep contextual embeddings, especially BERT, which captures sentence-level semantics. Models like Sentence-BERT~\cite{reimers2019sentencebert} further improve this by optimizing for semantic similarity tasks, enabling effective use in clustering and classification. These improvements are crucial, as studies show that the performance of clustering tasks depends more on the quality of embeddings than the choice of clustering algorithm itself~\cite{abdalgader2024transformer, subakti2022bert, petukhova2024llm}.

Multilingual capabilities have also been extended through knowledge distillation methods that adapt monolingual models into multilingual variants~\cite{reimers2020multilingual}. This is particularly relevant for global ITSM platforms. However, while embeddings allow for effective numeric representation of text, categorizing IT tickets via unsupervised clustering methods still presents challenges. Algorithms like K-means can be adapted to match historical category counts~\cite{aggarwal2013data}, but there is no guarantee that resulting clusters will align meaningfully with operational categories, especially when hundreds of labels are involved~\cite{suyal2014text}.

Applications of clustering and embedding techniques in other domains have shown promise, such as in scientific article classification~\cite{guedes2024contrastive}, topic modeling~\cite{george2023topic, susanto2023topic}, and ITSM recommender systems~\cite{reinhard2023recommender}. However, few studies have focused specifically on real-time categorization of IT tickets. For example,~\cite{costa2019itsm} reported poor accuracy using traditional ML for ticket categorization. While commercial tools like ServiceNow provide predictive features~(\cite{servicenow}), their internal methodologies remain undocumented.

Recent work on embedding-based clustering models also explores density-based methods for unsupervised categorization~\cite{radulescu2020density}, showing their feasibility but highlighting scalability and alignment challenges. Consequently, while significant advances have been made in representation learning and clustering, there is a clear need for domain-adapted solutions tailored to the structure and complexity of ITSM data.

From a performance perspective, even advanced pre-trained models like BERT or XLNet have shown limited effectiveness in IT ticket classification. ~\cite{ZANGARI2023} reported F1-scores between 38\% and 55\% on real-world datasets that is insufficient for deployment in production environments. The study attributed this to the domain gap between general-purpose language models and the technical jargon and formatting inconsistencies typical of IT tickets. Hierarchical classification, while promising in structured contexts, was less effective when applied to noisy or inconsistently labeled data.

Regarding evaluation, inter-rater agreement metrics such as Cohen’s Kappa~\cite{cohen1968} are commonly used to assess classification performance. Weighted variants of Cohen's Kappa are particularly useful for ordinal labels, such as severity or urgency levels. However, as noted by~\cite{datatab2024}, this metric is sensitive to class imbalance, an issue that is highly relevant in ITSM datasets, where medium-priority tickets vastly outnumber critical ones. This necessitates the use of complementary evaluation metrics and sampling strategies to ensure reliable model performance assessments.

In summary, while recent research highlights promising directions in the use of embeddings and deep learning for ticket classification and prioritization, existing approaches face significant limitations in scalability, multilingual handling, and domain adaptation. These gaps motivate the development of more robust and hybrid systems tailored to ITSM-specific challenges.

\section{Methods}
\label{sec:headings}

This section introduces the methods used in this study. We first formalize the prioritization task as a classification problem. 
We then present two main methods: an embedding-based approach that combines sentence-level representations with clustering and classical supervised methods (\ref{sec:methods:embedding}), and a fine-tuned transformer architecture designed to process textual and numerical features together (\ref{sec:methods:fine-tuning-transformer}).

\subsection{Problem formalization}
Ticket prioritization can be described as a standard classification problem where the target is a level of priority (integer). It enhances IT management efficiency by facilitating decision-making on which tickets should be prioritized, thereby minimizing downtime, and optimizes resource allocation, with critical incidents receiving immediate attention and resources. 
In this case, the priority is a function of the impact and urgency, shown in Table \ref{tab:priority_matrix}. For example, a high-urgency but low-impact issue can be described as having a moderate priority.

The whole model training pipeline is expected to run in real-time, ensuring scalability and cost efficiency, to handle incoming tickets to improve productivity.

Ticket data mainly consists of textual information, so we consider two approaches:
\begin{itemize}
    \item Embeddings with standard machine learning algorithm (\ref{sec:methods:embedding}),
    \item Fine-tuning a pre-trained transformer encoder on our task (\ref{sec:methods:fine-tuning-transformer}).
\end{itemize}

\begin{table}[ht]
\centering
\setlength{\tabcolsep}{10pt}
\renewcommand{\arraystretch}{1.3}

\begin{tabular}{lccc}
\toprule
 & \multicolumn{3}{c}{\textbf{Impact}} \\
\cmidrule(lr){2-4}
\textbf{Urgency} 
& \textbf{High (System Wide)} 
& \textbf{Medium (Multiple Users)} 
& \textbf{Low (Single User)} \\ 
\midrule

\textbf{High (Blocked)} 
& \cellcolor[RGB]{245,204,204} 1 -- Critical
& \cellcolor[RGB]{253,230,200} 2 -- High
& \cellcolor[RGB]{255,244,200} 3 -- Moderate
\\

\textbf{Medium (Impaired)} 
& \cellcolor[RGB]{253,230,200} 2 -- High
& \cellcolor[RGB]{255,244,200} 3 -- Moderate
& \cellcolor[RGB]{200,230,201} 4 -- Low
\\

\textbf{Low (Inconvenient)} 
& \cellcolor[RGB]{255,244,200} 3 -- Moderate
& \cellcolor[RGB]{200,230,201} 4 -- Low
& \cellcolor[RGB]{200,230,201} 4 -- Low
\\
\bottomrule
\end{tabular}

\caption{Priority matrix relating urgency and impact.}
\label{tab:priority_matrix}
\end{table}

\subsection{Embedding-based Approaches}\label{sec:methods:embedding}
This approach has the advantage of leveraging state-of-the-art pre-trained large language models (LLMs) that already learned complex language features at the sentence and corpus level, as well as word relationships, and with multilingual support. 
They are also cost-effective approaches because the language model is already trained, so its power footprint would mostly consist of memory and inference of new tickets. 

In particular, we use the sentence embedding model \texttt{sentence-transformers/paraphrase-multilingual-mpnet-base-v2} to convert input text to embedding vectors, as it has a competitive performance-to-size ratio. Then, we seek to apply standard machine learning classifiers in the embedding space, exploring both supervised and unsupervised approaches. 

\begin{figure}[htbp]
    \centering
    \includegraphics[width=\textwidth]{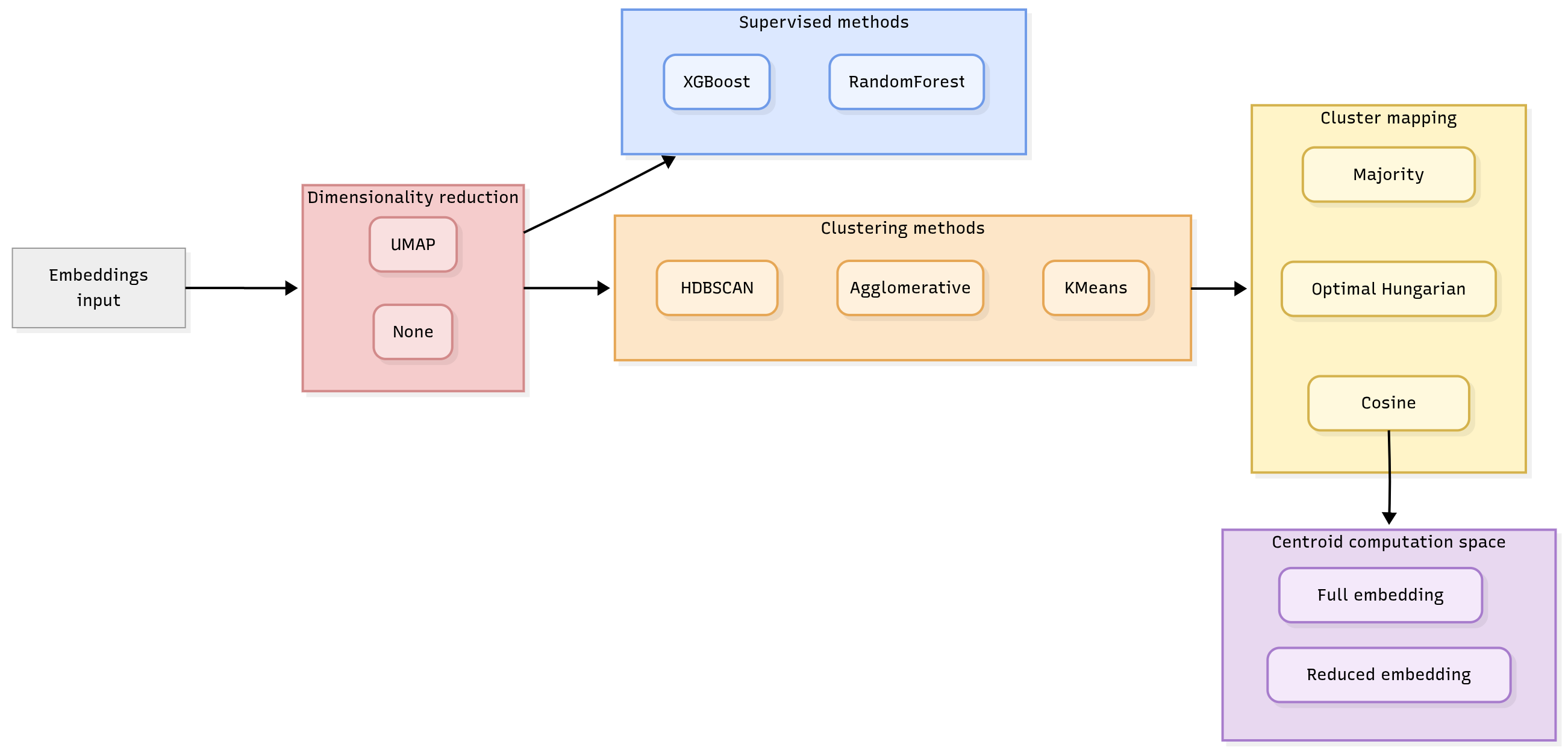} 
    \caption{Pipeline for embeddings-based approaches. Starting from embedding input, various modules can be combined: dimensionality reduction, clustering, supervised learning, cluster-mapping strategies, and centroid-space selection. Every final leaf node corresponds to a complete and executable pipeline.}
    \label{fig:mermaid-pipeline}
\end{figure}

Figure \ref{fig:mermaid-pipeline} shows an overview of the different pipelines considered for the experiments. This flexible framework allows for testing 30 different combinations.

However, the ability to test various machine learning methods also introduces challenges, as it requires careful selection and tuning to find the most effective approach for the data.
In addition, the performance depends on the quality of the pre-trained model, which has not been specialized in IT language. So if the sentence embedding inputs are low-quality, then the subsequent clustering or classification model will struggle to perform properly. 
Moreover, it cannot process both text and other numerical features (e.g., time intervals). 

Plus, clustering and decision tree-based methods, such as XGBoost, are not suitable to handle the output of two variables.
To overcome this issue, we can define a combinatorial output variable that varies from 0 to 15, since there are four possible levels of urgency and impact, as shown in Eq. \ref{eq:combined_output}.
\begin{equation}
\label{eq:combined_output}
\mathrm{Combined\_output} = 4*\,\mathrm{urgency} + \mathrm{impact}
\end{equation}


The following sections present each module and its respective roles within the pipeline, detailing the available configuration choices.

\subsubsection{Dimensionality reduction}
The embedding space resulting from \texttt{paraphrase-multilingual-mpnet-base-v2} is of size 768, so we can either reduce its dimensions to simplify upstream clustering or supervised learning, or keep the original embedding space.

To reduce the input dimensions, we can use UMAP \citep{mcinnes2020umapuniformmanifoldapproximation}, which is compatible with non-linear inputs. It can help the downstream clustering algorithm by making clusters more compact and separated in lower dimensions. It can also scale to large datasets.
For UMAP, the main hyperparameter of interest is the target output dimensions.


\subsubsection{Clustering}

After the previous step, the embedded data can be clustered.
Clustering assumes that tickets of the same impact and urgency are similar and can form a coherent cluster.
We selected several clustering algorithms:
\begin{itemize}
    \item HDBSCAN \citep{hdbscan_Campello_2013},
    \item Agglomerative clustering \citep{mullner2011modern},
    \item K-Means \citep{lloyd1982least}.
\end{itemize}

HDBSCAN is density-based and hierarchical; it has the advantages of being able to cluster of different densities and detecting outliers, so it is well-suited for high dimensions.

Agglomerative clustering is bottom-up and hierarchical, where each sample starts as its own cluster, and they are merged iteratively based on a distance criterion. Thus, it can exploit very flexible structures without assumptions, but is sensitive to its criterion.

K-Means is a centroid-based algorithm that partitions data into a predefined number of clusters and iteratively assigning points to the nearest centroid, and then updates all centroids. It is computationally efficient, but is not robust to non-linear cluster structures or outliers.

We apply these clustering methods on the embedded input space, resulting in a set of unlabeled clusters that need to be classified.

\subsubsection{Cluster mapping}

After clustering, we obtain a set of unlabeled clusters that needs to be classified.

In this effect, we found three main strategies to assign clusters:
\begin{itemize}
    \item Majority-class assignment,
    \item Optimal Hungarian algorithm \citep{kuhn1955hungarian},
    \item Cosine-similarity based.
\end{itemize}

The majority-class assignment uses the most frequent label present in each newfound cluster, but is sensitive to class imbalance. 
It can be used both with and without the clustering module.

The Hungarian algorithm was designed to construct a complete assignment of workers (clusters in our case) to jobs (true labels) at a minimal cost, where each assignment has a defined cost (error).

We can also assign clusters based on cosine similarity to centroids, where tickets are assigned to the cluster whose centroid is closest to their embedding vector.
In this case, we can use the embedding vector from either the reduced space produced by UMAP or the original embedding space (mentioned in Fig. \ref{fig:mermaid-pipeline}).

\subsubsection{Supervised classification}

We experiment with two widely used supervised models:

\begin{itemize}
    \item XGBoost \citep{chen2016xgboost},
    \item Random Forest \citep{breiman2001randomforest}.
\end{itemize}

In addition to clustering-based approaches, we train both models to predict the combined target label based on the embedded representations. These supervised baselines enable us to evaluate how well traditional machine learning methods can leverage the learned embeddings, serving.

\subsection{Fine-tuning a custom transformer-based architecture}\label{sec:methods:fine-tuning-transformer}

\begin{figure}[htbp]  
    \centering
    \includegraphics[width=\textwidth]{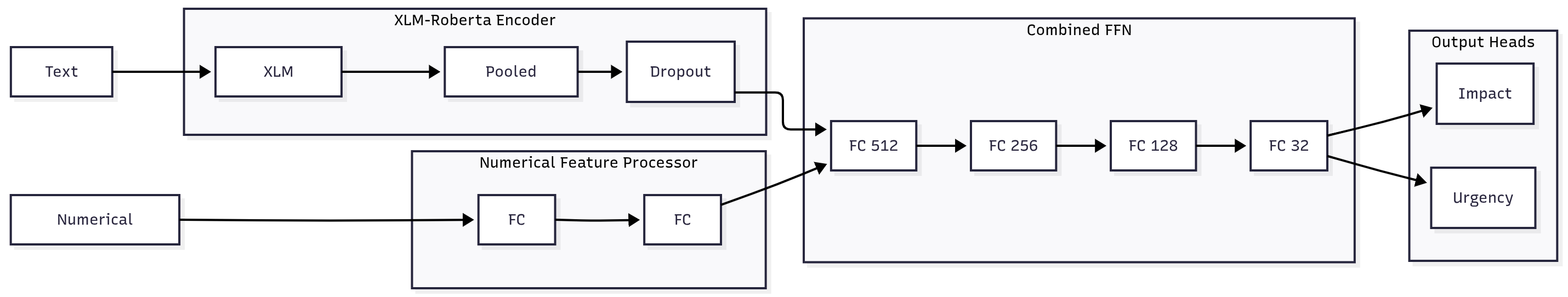} 
    \caption{Architecture of the fine-tuned transformer encoder model.}
    \label{fig:xlm-roberta-custom}
\end{figure}

We also design a supervised architecture based on a multilingual transformer encoder, as shown in Fig. \ref{fig:xlm-roberta-custom}.
The model integrates a pre-trained transformer encoder \citep{conneau2019_xlm} backbone (\texttt{FacebookAI/xlm-roberta-base}), and in parallel, it features a numerical feature processor and a shared feed-forward network (FFN), enabling joint exploitation of textual and numerical information for predicting impact and urgency. The full architecture has approximately 278M parameters.

Textual input is encoded by \texttt{XLM-RoBERTa} and we extract the pooled representation, while numerical features are transformed by a dedicated FFN. 
The two representations are then concatenated and passed through a four-layer FFN with PReLU activations, Then, this shared representation is mapped to the two targets using separate linear heads in a multi-output setting.

The advantages of this approach are that it leverages state-of-the-art pre-trained models that are specifically trained to encode complex multilingual contexts while providing the flexibility of the deep learning paradigm by designing custom architectures on top of the pre-trained LLMs and thus allowing the processing of numerical features in parallel, as well as making simultaneous predictions for the impact and urgency variables. Moreover, deep learning generally provides more capabilities to capture very complex features and structures from data. It can also learn to specialize in IT tickets.

The disadvantages of this approach are the same as deep learning: high computational and time costs for training the model and making inferences. Additionally, implementing and fine-tuning LLMs is complex and requires expertise in both deep learning and domain-specific data. Some common issues that prevent obtaining good models are a lack of training data and overfitting. Plus, they are black box models, making it difficult to interpret what they did learn or not, as well as their decisions.


\section{Experimental Setup}
\label{sec:experimental-setup}

In this section, we detail the experimental setup used to evaluate our approach. We first present the dataset and preprocessing steps (\ref{sec:dataset}). Next, we define the evaluation metrics used to quantify model performance for this application (\ref{sec:metrics}). We then introduce the confidence score computation (\ref{sec:confidence_score}), which provides a quantitative measure of the confidence of individual predictions. Finally, we describe the training setup for the transformer model, including key hyperparameter choices (\ref{sec:training_config}).

\subsection{Dataset}\label{sec:dataset}

Our dataset comprises approximately 26,000 real industrial IT service tickets over one year, including both textual descriptions and numerical fields.

Our preprocessing applies a small set of targeted cleaning steps. We remove exact duplicates, repeated ticket IDs, and obvious noise such as test tickets, then discard rows with missing values. HTML fields are converted to plain text by stripping non-informative elements, and the text is normalized by removing residual special characters and converting to lowercase.

We use the following fields for each ticket :
\begin{itemize}
    \item Title,
    \item Description,
    \item Category,
    \item Department,
    \item Related asset name,
    \item Description of related asset,
    \item Submit date
    \item Max resolution date.
\end{itemize}
Text fields are concatenated together with a prefix prompt (e.g., \texttt{Ticket title: \textit{<title>}})
The numerical feature contains a value of the time interval between the max resolution date and the submit date, providing information on the complexity and urgency of the ticket. This value is scaled using a min-max scaler fitted on the train set and applied to the validation and test sets.

A central challenge of this dataset is the strong class imbalance present in both target variables impact and urgency which exhibit highly skewed distributions with one dominant class and several rare ones (\ref{tab:dataset_distribution}). The task also requires predicting both targets jointly, introducing dependencies between them, and increasing the risk of overfitting to majority patterns.

For the split, we use a stratified split on urgency labels with 60\%/20\%/20\%.

\begin{table}[h]
\centering
\begin{tabular}{c|r|r||c|r|r}
\multicolumn{3}{c||}{\textbf{Urgency}} &
\multicolumn{3}{c}{\textbf{Impact}} \\
\hline
\textbf{Label} & \textbf{Count} & \textbf{Percentage} &
\textbf{Label} & \textbf{Count} & \textbf{Percentage} \\
\hline
1 & 1075  & 4.01\% &
1 & 428   & 1.60\% \\
2 & 5745  & 21.43\% &
2 & 168   & 0.63\% \\
3 & 8385  & 31.27\% &
3 & 25392 & 94.70\% \\
4 & 1276  & 4.76\% &
4 & 825   & 3.08\% \\
5 & 10332 & 38.53\% &
\multicolumn{3}{c}{} \\
\end{tabular}
\label{tab:dataset_distribution}
\caption{Label distributions for urgency and impact over the full dataset.}
\end{table}

\subsection{Evaluation metrics}\label{sec:metrics}

We evaluate model performance using standard classification metrics, including F1-score and Cohen's Kappa \citep{cohen1968}.  
Cohen's Kappa is sensitive to class imbalance, which is high in our case (\ref{tab:dataset_distribution}), so we choose to focus on the average F1-score of impact and urgency.

For scalability, we also measure inference time, computational cost, and RAM usage.

\subsection{Confidence score}\label{sec:confidence_score}

A key requirement of our application is to output, alongside each prediction, a confidence score that is easy to interpret and does not require additional training or architectural changes. We therefore seek a score that (i) is expressed as a percentage starting from zero and (ii) can be computed directly from the model’s logits.

In particular, a sign of an uncertain prediction could be close values between the first largest and second largest value.
To measure this effect, we define a confidence margin-based metric between the highest logit $y_{\text{top1}}$ and the second-highest $y_{\text{top2}}$. This margin can be seen as a confidence gap, normalized by the sum of their absolute values and converted into a percentage, as defined in Eq. \ref{eq:confidence}. It indicates how much more confident the model is in its top choice compared to the next best option.

\begin{equation}\label{eq:confidence}
\text{Confidence}(\hat{y}) = 
\left( 
\frac{y_{\text{top1}} - y_{\text{top2}}}{\lvert y_{\text{top1}} \rvert + \lvert y_{\text{top2}} \rvert} 
\right) \times 100
\end{equation}

\subsection{Training configuration}\label{sec:training_config}

For the fine-tuned transformer model, we use a NVIDIA A10G GPU. Training time is set to six hours, and we select the best model based on the average F1-score of impact and urgency on the validation set. 
The hyperparameter experiments focused mainly on the FC layer dimensions, dropout rates, and batch sizes.

\section{Results}
\label{sec:results}

In this section, we compare the experimental results, first for embedding-based approaches (\ref{sec:res:embedding}) and then for the fine-tuned transformer architecture (\ref{sec:res:finetuning}).

\subsection{Embedding-based Approaches}\label{sec:res:embedding}

\subsubsection{Train results}

We report our extensive experiments with clustering-based methods in Table \ref{tab:train_embedding_results}, and supervised methods in Table \ref{tab:train_supervised_results}.
Clustering-based methods perform poorly on the train set, and therefore we do not evaluate them on validation or test data. We also conclude that the choice of clustering algorithm has little effect on the outcomes. 
The best result achieved is 39.96\% F1-score on Urgency, using UMAP with Agglomerative Clustering and majority-class assignment. 
In contrast, a random classifier yields only 3.40\% F1-score, and the worst performance 0.69\% F1-score overall is obtained with UMAP and HDBSCAN using cosine similarity in the reduced space. Overall, these results confirm that clustering-based baselines are largely ineffective for this task.

\begin{table}[h!]
\centering
\caption{Train set results for embedded-based approaches with unsupervised methods}\label{tab:train_embedding_results}
\resizebox{\textwidth}{!}{%
\begin{tabular}{lllcccccc}
\hline
\textbf{Dim. Reduction} & \textbf{Clustering} & \textbf{Assignment / Scoring} & \textbf{Combined Acc \%} & \textbf{Combined F1} & \textbf{Accuracy Urgency} & \textbf{F1 Urgency} & \textbf{Accuracy Impact} & \textbf{F1 Impact} \\
\hline
None & None & Random & 7.21 & 3.4 & 26.08 & 23.11 & 28.55 & 13.81 \\
UMAP & HDBSCAN & Majority & 40.79 & 4.85 & 44.38 & 17.93 & \textbf{94.41} & 24.28 \\
UMAP & HDBSCAN & Optimal Hungarian & 34.1 & 5.28 & 39.87 & 16.98 & 89.87 & \textbf{24.47} \\
UMAP & HDBSCAN & Cosine sim, full & 3.61 & 1.22 & 32.28 & 13.95 & 7.27 & 4.27 \\
UMAP & HDBSCAN & Cosine sim, reduced & 3.53 & 0.69 & 23.73 & 10.43 & 19.25 & 9.19 \\
UMAP & AggClustering & Majority & 54.62 & 11 & 55.05 & \textbf{39.96} & \textbf{94.41} & 24.28 \\
UMAP & AggClustering & Optimal Hungarian & 30.63 & 10.92 & 39.41 & 32.73 & 55.16 & 21.77 \\
UMAP & AggClustering & Cosine sim, full & 6.16 & 3.24 & 13.29 & 12.67 & 31.42 & 14.23 \\
UMAP & AggClustering & Cosine sim, reduced & 6.38 & 2.87 & 18.06 & 15.87 & 42.11 & 17.15 \\
UMAP & KMeans & Majority & \textbf{55.01} & 10.05 & 55.3 & 34.36 & \textbf{94.41} & 24.28 \\
UMAP & KMeans & Optimal Hungarian & 27.13 & 9.82 & 36.87 & 30.78 & 50.85 & 20.74 \\
UMAP & KMeans & Cosine sim, full & 2.45 & 1.24 & 20.16 & 15.97 & 18.7 & 9.52 \\
UMAP & KMeans & Cosine sim, reduced & 3.51 & 1.35 & 15.81 & 14.65 & 21.2 & 10.1 \\
UMAP & Direct clustering & Reduced space & 21.9 & 8.15 & 35.85 & 28.66 & 37.4 & 18.47 \\
None & HDBSCAN & Majority & 46.93 & 7.95 & 47.52 & 27.51 & \textbf{94.41} & 24.28 \\
None & HDBSCAN & Optimal Hungarian & 0.5 & 0.19 & 24.56 & 16.16 & 1.28 & 1.19 \\
None & HDBSCAN & Cosine sim, full space & 37.37 & 4.48 & 43.57 & 19.98 & 84.47 & 23.67 \\
None & AggClustering & Majority & 51.41 & 9.82 & 52.23 & 34.06 & \textbf{94.41} & 24.28 \\
None & AggClustering & Optimal Hungarian & 25.55 & 9.87 & 35.21 & 30.12 & 48.12 & 20.54 \\
None & AggClustering & Cosine sim, full & 7.44 & 3.29 & 36.36 & 31.08 & 25.9 & 11.79 \\
None & KMeans & Majority & 54.49 & \textbf{11.07} & \textbf{56.39} & 39.04 & \textbf{94.41} & 24.28 \\
None & KMeans & Optimal Hungarian & 22.32 & 9.64 & 45.23 & 38.86 & 38.26 & 18.51 \\
None & KMeans & Cosine sim, full & 4.69 & 2.89 & 33.21 & 27.13 & 27.74 & 14.59 \\
\hline
\end{tabular}%
}
\end{table}

We vary the number of estimators in XGBOOST and Random Forest, and report results in Table \ref{tab:train_supervised_results}.

\begin{table*}[t]
\centering
\caption{Train set results for supervised embedded-based approaches}\label{tab:train_supervised_results}
\resizebox{\textwidth}{!}{%
\begin{tabular}{lllcccccc}
\toprule
\textbf{Dim. Reduction} &
\textbf{Model} &
\textbf{Estimators} &
\textbf{Combined Acc \%} &
\textbf{Combined F1} &
\textbf{Accuracy Urgency} &
\textbf{F1 Urgency} &
\textbf{Accuracy Impact} &
\textbf{F1 Impact} \\
\midrule
\multicolumn{9}{c}{\textbf{UMAP + XGBoost}} \\
\midrule
UMAP & XGBoost & 100 & 78.48 & 71.08 & 79.28 & 76.22 & 97.42 & 82.18 \\
UMAP & XGBoost & 150 & 86.28 & 79.38 & 86.63 & 86.30 & 98.84 & 94.09 \\
UMAP & XGBoost & 200 & 90.62 & 96.20 & 90.78 & 90.64 & 99.43 & 97.32 \\
UMAP & XGBoost & 250 & 96.16 & 98.18 & 94.21 & 94.74 & 99.83 & 99.33 \\
UMAP & XGBoost & 300--500 & 96.75--\textbf{99.62} & 91.94--99.91 & 96.76--\textbf{99.62} & 97.12--99.70 & 99.96--\textbf{100.00} & 99.77--\textbf{100.00} \\
\midrule
\multicolumn{9}{c}{\textbf{UMAP + Random Forest}} \\
\midrule
UMAP & Random Forest & 100--500 & 99.99--\textbf{100.00} & 99.99--\textbf{100.00} & \textbf{100.00} & \textbf{100.00} & 99.99--\textbf{100.00} & \textbf{100.00} \\
\midrule
\multicolumn{9}{c}{\textbf{XGBoost (no Dim. Reduction)}} \\
\midrule
None & XGBoost & 100 & 97.35 & 92.10 & 97.35 & 97.35 & 99.98 & 99.79 \\
None & XGBoost & 150--500 & 99.28--99.97 & 99.82--99.99 & 99.28--99.97 & 99.40--99.97 & 99.99--\textbf{100.00} & 99.98--\textbf{100.00} \\
\midrule
\multicolumn{9}{c}{\textbf{Random Forest (no Dim. Reduction)}} \\
\midrule
None & Random Forest & 100--500 & 99.99--\textbf{100.00} & 99.99--\textbf{100.00} & \textbf{100.00} & \textbf{100.00} & 99.99--\textbf{100.00} & \textbf{100.00} \\
\bottomrule
\end{tabular}
}
\end{table*}

From the train results, we observe that Random Forest and XGBOOST are clearly overfitting, but keep them for the validation test round.

\subsubsection{Validation results}

Here we report the validation result with embedding-based methods with XGBOOST and Random Forest in Table \ref{tab:val_embedding_results}
We find that the best model on averaged F1-score is UMAP with XGBOOST with 350 estimators with 24.05\% F1-score overall, 54.28\% F1-score on urgency, 45.54\% F1-score on impact overall.

\begin{table}[th]
\centering
\caption{Validation set results for supervised embedded-based approaches}\label{tab:val_embedding_results}
\resizebox{\textwidth}{!}{%
\begin{tabular}{lllcccccc}
\hline
\textbf{Dim. Reduction} & \textbf{Model} & \textbf{Estimators} & \textbf{Combined Acc \%} & \textbf{Combined F1} & \textbf{Accuracy Urgency} & \textbf{F1 Urgency} & \textbf{Accuracy Impact} & \textbf{F1 Impact} \\
\hline
UMAP & XGBOOST & 100 & 60.06 & 21.73 & 62.16 & 53.86 & 93.93 & 41.3 \\
UMAP & XGBOOST & 150 & 60.11 & 23.59 & 62.37 & 54.97 & 94.22 & 44.34 \\
UMAP & XGBOOST & 200 & 59.22 & 20.54 & 61.55 & 53.41 & 93.9 & 42.04 \\
UMAP & XGBOOST & 250 & 59.51 & 19.93 & 61.76 & 53.21 & 93.81 & 40.92 \\
UMAP & XGBOOST & 300 & 59.4 & 22.48 & 61.67 & 53.35 & 93.95 & 42.58 \\
UMAP & XGBOOST & 350 & 59.63 & \textbf{24.05} & 61.83 & 54.28 & 94.16 & 45.54 \\
UMAP & XGBOOST & 400 & 59.36 & 22.08 & 61.64 & 53.31 & 94.22 & 44.6 \\
UMAP & XGBOOST & 450 & 58.84 & 21.18 & 61.17 & 52.57 & 94.06 & 42.73 \\
UMAP & XGBOOST & 500 & 59.85 & 23.5 & 61.96 & 53.31 & 94.16 & 43.86 \\
UMAP & Random Forest & 100 & 57.34 & 20.19 & 58.99 & 49.34 & 94.04 & 38.75 \\
UMAP & Random Forest & 150 & 57.64 & 19.57 & 59.61 & 50.59 & 94.07 & 39.24 \\
UMAP & Random Forest & 200 & 58.11 & 17.97 & 60.2 & 51.12 & 93.54 & 36.76 \\
UMAP & Random Forest & 250 & 57.07 & 20.78 & 58.84 & 49.34 & 93.86 & 39.99 \\
UMAP & Random Forest & 300 & 57.82 & 20.67 & 59.61 & 49.14 & 93.86 & 39.15 \\
UMAP & Random Forest & 350 & 58.15 & 21.93 & 60 & 51.31 & 93.88 & 41.42 \\
UMAP & Random Forest & 400 & 56.96 & 18.95 & 85.97 & 49.98 & 93.75 & 39.72 \\
UMAP & Random Forest & 450 & 57.45 & 17.82 & 59.58 & \textbf{59.17} & 93.63 & 36.58 \\
UMAP & Random Forest & 500 & 58.06 & 19.85 & 60.03 & 50.16 & 93.72 & 37.88 \\
None & XGBOOST & 100 & 62.8 & 22.98 & 64.93 & 54.5 & 95.15 & 42.3 \\
None & XGBOOST & 150 & 63.41 & 23.57 & 65.54 & 55.44 & 95.29 & 43.93 \\
None & XGBOOST & 200 & 63.3 & 23.42 & 65.43 & 55.14 & 95.26 & 43.66 \\
None & XGBOOST & 250 & 63.62 & 23.42 & 65.75 & 55.11 & 95.26 & 43.66 \\
None & XGBOOST & 300 & 63.68 & 23.64 & 65.88 & 55.49 & 95.31 & 44.31 \\
None & XGBOOST & 350 & 63.59 & 23.57 & 65.75 & 55.49 & 95.27 & 44.08 \\
None & XGBOOST & 400 & \textbf{64.05} & 24.01 & \textbf{66.24} & 56.07 & 95.36 & 45.1 \\
None & XGBOOST & 450 & 64.02 & 24.02 & 66.18 & 56.06 & \textbf{95.38} & 45.14 \\
None & XGBOOST & 500 & 64.02 & 23.96 & 66.2 & 56.2 & 95.31 & 44.94 \\
None & Random Forest & 100 & 61.49 & 16.43 & 61.27 & 46.48 & 94.81 & 29.44 \\
None & Random Forest & 150 & 61.85 & 16.34 & 63.64 & 46.79 & 94.77 & 28.69 \\
None & Random Forest & 200 & 61.76 & 16.35 & 63.53 & 46.76 & 94.77 & 28.93 \\
None & Random Forest & 250 & 62.08 & 16.31 & 63.82 & 46.68 & 94.75 & 28.49 \\
None & Random Forest & 300 & 62.1 & 16.56 & 63.71 & 46.73 & 94.81 & 29.25 \\
None & Random Forest & 350 & 61.94 & 16.42 & 63.59 & 46.65 & 94.79 & 28.97 \\
None & Random Forest & 400 & 61.85 & 16.54 & 63.69 & 56.82 & 94.81 & 29.44 \\
None & Random Forest & 450 & 61.82 & 16.43 & 63.53 & 46.71 & 94.79 & 28.97 \\
None & Random Forest & 500 & 61.08 & 16.02 & 62.8 & 45.82 & 94.75 & 28.49 \\
\hline
\end{tabular}%
}
\end{table}


\subsubsection{Test results}

Based on the results on the val set, selecting UMAP with XGBOOST with 350 estimators as the best model, the final test results are 23.7\% F1-score overall, 52.16\% F1-score on urgency, and 42.57\% on impact.

\subsubsection{Conclusion on embedding-based approaches.}

We conclude that the embedding based classification approach is not suitable for the prioritization problem because of very poor results.

There are several potential reasons for these results.

\paragraph{Dataset quality.}

The features of the dataset consist of human-written texts with inherent noise and variability in writing style, typographical as well as grammatical errors, making it challenging to process and analyze effectively. It can also contain very specific or complex IT domain knowledge.

\paragraph{UMAP and Clustering.}

The UMAP documentation \citep{umap_learn_docs_clustering} warns that applying clustering on UMAP (or t-SNE) embeddings can produce misleading clusters that appear meaningful in plots. Apparent clusters may reflect distortions in the embedding rather than true semantic groupings, making it difficult to know whether the detected structure corresponds to meaningful patterns in the data or just the shape of the point cloud.

\paragraph{Embedding quality.}

We also suspect that the embedding outputs from `sentence-transformers/paraphrase-multilingual-mpnet-base-v2` are not fitted to ITSM ticket data, and are a more general LLMs. Thus, this greatly impacts the embedding output quality of the dataset and may result in lower quality inputs for the clustering models.

In the case of supervised tree-based methods, the lower quality of inputs can hinder the training and generalization phase, resulting in poor classification.

\subsection{Fine-tuning a custom transformer-based architecture}\label{sec:res:finetuning}

The test set results (Table~\ref{tab:results:xlm}) confirm that the model performs well overall, with some expected limitations due to class imbalance. Accuracy is inflated by this imbalance, whereas weighted Cohen's kappa remains high (\(\approx 0.80\)), indicating that predictions largely align with the true labels and that most errors are close to the correct class. Although kappa is still sensitive to imbalance, the weighting reduces the impact of small deviations, making it a more reliable metric than raw accuracy.  

The F1-score for the impact variable reaches 86.80\%, which is strong, while the urgency F1-score is lower at 70.22\%, yet still reasonable given the noisier and more subjective nature of this target.  

To better interpret these results, we examine the confusion matrices. For impact (Table~\ref{tab:conf_mat:impact}), predictions are mostly accurate, consistent with the high F1 and kappa scores. The main errors arise from confusion between classes~0 and~2, slightly reducing agreement and explaining part of the kappa degradation. Nevertheless, the model effectively captures the structure of impactful tickets.  

For urgency (Table~\ref{tab:conf_mat:urgency}), performance is lower, with the largest confusion between classes~1 and~2 and unreliable predictions for class~3 due to limited samples. These patterns correspond to the observed lower F1-score for this target. 

Overall, the performance gap between the two variables may be reflected by their nature: impact is more objective, driven by technical factors that provide clear patterns, whereas urgency is more subjective and context-dependent, influenced by user behavior and situational factors, which increases variability and makes prediction inherently more challenging.

\begin{table}[h!]
\centering
\caption{Confusion matrix for Impact on the test set}\label{tab:conf_mat:impact}
\begin{tabular}{c|cccc}
\toprule
\textbf{True / Pred} & \textbf{0} & \textbf{1} & \textbf{2} & \textbf{3} \\
\midrule
\textbf{0} & 97  & 0  & 8    & 0   \\
\textbf{1} & 1   & 26 & 17   & 0   \\
\textbf{2} & 38  & 3  & 5313 & 16  \\
\textbf{3} & 0   & 0  & 19   & 179 \\
\bottomrule
\end{tabular}
\end{table}

\begin{table}[h!]
\centering
\caption{Confusion matrix for Urgency on the test set}\label{tab:conf_mat:urgency}
\begin{tabular}{c|ccccc}
\toprule
\textbf{True / Pred} & \textbf{0} & \textbf{1} & \textbf{2} & \textbf{3} & \textbf{4} \\
\midrule
\textbf{0} & 147 & 69  & 13   & 2   & 5   \\
\textbf{1} & 26  & 804 & 360  & 7   & 41  \\
\textbf{2} & 8   & 471 & 1180 & 13  & 93  \\
\textbf{3} & 3   & 53  & 127  & 27  & 68  \\
\textbf{4} & 0   & 15  & 51   & 18  & 2116 \\
\bottomrule
\end{tabular}
\end{table}

\begin{table}[h!]
\centering
\caption{Test set evaluation}\label{tab:results:xlm}
\begin{tabular}{l c}
\toprule
\textbf{Metric} & \textbf{Value} \\
\midrule
Accuracy (Impact)                & 0.9830 \\
Accuracy (Urgency)               & 0.7653 \\
F1-score (Impact)                & 0.8680 \\
F1-score (Urgency)               & 0.7022 \\
\textbf{F1-score (Average)}      & \textbf{0.7851} \\
Weighted Cohen's Kappa (Impact)  & 0.8236 \\
Weighted Cohen's Kappa (Urgency) & 0.8624 \\
Weighted Linear Kappa (Impact)   & 0.8392 \\
Weighted Linear Kappa (Urgency)  & 0.7869 \\
\bottomrule
\end{tabular}
\end{table}

\paragraph{Scaling tests.} 
After training, the model requires 5.33GB of CPU RAM, and can perform a single inference in 0.8 seconds on CPU only, which is scalable for real-time inference deployment.

\section*{Conclusion and future work}
\label{sec:conclusion}

This work compared embedding-based pipelines with a fine-tuned transformer for predicting the impact and urgency of ITSM tickets. Embedding-based methods, although efficient, were limited by the performance of the quality of the dataset, the constraints introduced by dimensionality reduction and clustering, and the poor IT-domain representation of the embedding. The fine-tuned transformer delivered consistently stronger performance, effectively leveraging textual representations and numerical features to predict IT ticket priorities. Remaining gaps, particularly the lower performance on urgency, largely reflect the dataset’s subjectivity and class imbalance. 

Future work could first explore more specialized prediction heads for each target, aiming to reduce the performance gap between impact and urgency. Another direction is to combine the fine-tuned transformer with sentence-embedding-based representations to enrich the feature space.

\bibliographystyle{plainnat}

\end{document}